\documentclass[journal,twoside]{IEEEtran}

\usepackage{color}
\usepackage{algorithm}
\usepackage{multirow}

\usepackage{hyperref}
\hypersetup{
  colorlinks   = true, 
  urlcolor     = blue, 
  linkcolor    = blue, 
  citecolor   = red 
}

\usepackage[noadjust]{cite}

\ifCLASSINFOpdf
   \usepackage[pdftex]{graphicx}
  \graphicspath{{img/}}
  \DeclareGraphicsExtensions{.pdf, .png, .jpg}
\else
\fi
\usepackage{amsmath}
\usepackage{amssymb}
\usepackage{algorithmic}

\usepackage{array}

\ifCLASSOPTIONcompsoc
  \usepackage[caption=false,font=normalsize,labelfont=sf,textfont=sf]{subfig}
\else
  \usepackage[caption=false,font=footnotesize]{subfig}
\fi

\usepackage{stfloats}
\ifCLASSOPTIONcaptionsoff
  \usepackage[nomarkers]{endfloat}
 \let\MYoriglatexcaption\caption
 \renewcommand{\caption}[2][\relax]{\MYoriglatexcaption[#2]{#2}}
\fi
\usepackage{url}

\begin{document}
%
\title{DeepSleepNet: a Model for Automatic Sleep Stage Scoring based on Raw Single-Channel EEG}
%
%
%

\author{
Akara~Supratak,~Hao~Dong,~Chao~Wu,~and~Yike~Guo\text{*}
\thanks{A.~Supratak, H.~Dong, C.~Wu and Y.~Guo are with the Department of Computing, Imperial College London, London, SW7 2AZ, UK (e-mail: \{as12212, hao.dong11, chao.wu, y.guo\}@ic.ac.uk)}
\thanks{\text{*} Corresponding author}%
\thanks{DOI: \href{http://ieeexplore.ieee.org/document/7961240/}{10.1109/TNSRE.2017.2721116}}%
\thanks{Copyright has been transferred to IEEE}%
}

%
%

\markboth{This article has been published in IEEE Transactions on Neural Systems and Rehabilitation Engineering.}%
{Supratak \MakeLowercase{\textit{et al.}}: DeepSleepNet: a Model for Automatic Sleep Stage Scoring based on Raw Single-Channel EEG}
%



\maketitle

\begin{abstract}
The present study proposes a deep learning model, named DeepSleepNet, for automatic sleep stage scoring based on raw single-channel EEG. Most of the existing methods rely on hand-engineered features which require prior knowledge of sleep analysis. Only a few of them encode the temporal information such as transition rules, which is important for identifying the next sleep stages, into the extracted features. In the proposed model, we utilize Convolutional Neural Networks to extract time-invariant features, and bidirectional-Long Short-Term Memory to learn transition rules among sleep stages automatically from EEG epochs.
We implement a two-step training algorithm to train our model efficiently.
We evaluated our model using different single-channel EEGs (F4-EOG(Left), Fpz-Cz and Pz-Oz) from two public sleep datasets, that have different properties (e.g., sampling rate) and scoring standards (AASM and R\&K). The results showed that our model achieved similar overall accuracy and macro F1-score (MASS: 86.2\%-81.7, Sleep-EDF: 82.0\%-76.9) compared to the state-of-the-art methods (MASS: 85.9\%-80.5, Sleep-EDF: 78.9\%-73.7) on both datasets. This demonstrated that, without changing the model architecture and the training algorithm, our model could automatically learn features for sleep stage scoring from different raw single-channel EEGs from different datasets without utilizing any hand-engineered features. Our code is publicly available at \url{https://github.com/akaraspt/deepsleepnet}. The final version of this paper can be found in \url{http://ieeexplore.ieee.org/document/7961240/}.
\end{abstract}


%
\IEEEpeerreviewmaketitle

\section{Introduction}
%
%
%
%
%

\IEEEPARstart{S}{leep} plays an important role in human health. Being able to monitor how well people sleep has a significant impact on medical research and practice~\cite{Wulff2010}. 

Typically, sleep experts determine the quality of sleep using electrical activity recorded from sensors attached to different parts of the body. A set of signals from these sensors is called a polysomnogram (PSG), consisting of an electroencephalogram (EEG), an electrooculogram (EOG), an electromyogram (EMG), and an electrocardiogram (ECG). This PSG is segmented into \mbox{30-s} epochs, which are then be classified into different sleep stages by the experts according to sleep manuals such as the Rechtschaffen and Kales (R\&K)~\cite{AllanHobson1969} and the American Academy of Sleep Medicine (AASM)~\cite{iber2007}. This process is called sleep stage scoring or sleep stage classification. This manual approach is, however, labor-intensive and time-consuming due to the need for PSG recordings from several sensors attached to subjects over several nights.

There have been a number of studies trying to develop a method to automate sleep stage scoring based on multiple signals such as EEG, EOG and EMG~\cite{lajnef2015,huang2014,gunes2010}, or single-channel EEG~\cite{tsinalis2016,sharma2017,hassan2017}. These methods firstly extract time-domain, frequency-domain and time-frequency-domain features from each recording epoch. 
In the case of multiple signals, the features from all signals in one epoch were concatenated into one feature vector.
The features are then used to train classifiers to identify the sleep stage of the epoch. However, we believe that these methods may well not generalize to a larger population due to the heterogeneity among subjects and recording hardware. This is because these features were hand-engineered based on the characteristics of the available dataset.

Recently, deep learning, a branch of machine learning that utilizes multiple layers of linear and non-linear processing units to learn hierarchical representations or features from input data, has been employed in sleep stage scoring. For instance, the authors in \cite{langkvist2012} have investigated a capability of Deep Belief Nets (DBNs) to learn probabilistic representations from preprocessed raw PSG. Convolutional Neural Networks (CNNs) have also been applied to learn multiple filters that are used to convolve with small portions of input data (i.e., convolution) to extract time-invariant features from raw Fpz-Cz EEG channel~\cite{tsinalis2016cnn}. However, the results from the literature showed that applying deep learning on hand-engineered features performed better than on raw signals~\cite{langkvist2012,tsinalis2016}. This might well be because the authors did not consider temporal information that sleep experts use when they determine the sleep stage of each epoch.

Only a few number of literature have explored Recurrent Neural Networks (RNNs) in sleep stage scoring. RNNs are capable of conditioning outputs on all previous inputs, as they maintain internal memory and utilizes feedback (or loop) connections to learn temporal information from sequences of inputs. The main advantage of RNNs is that they can be trained to learn long-term dependencies such as transition rules~\cite{iber2007} that sleep experts use to identify the next possible sleep stages from a sequence of PSG epochs. Elman RNNs have been applied on energy features from the Fpz-Cz EEG channel~\cite{hsu2013}. In our previous work~\cite{dong2016lstm}, we also applied Long Short-Term Memory (LSTM) on time-frequency-domain features from the F4-EOG and Fp2-EOG channels separately. Even though the reported results were promising, these methods still rely on hand-engineered features.

This paper introduces \textit{DeepSleepNet}, a model for automatic sleep stage scoring based on raw single-channel EEG, which is different from the existing works that develop algorithms to extract features from EEG. We aim to automate the process of hand-engineering features by utilizing the feature extraction capabilities of deep learning. The main contributions of this work are as follows:
\begin{itemize}
  \item We develop a new model architecture that utilizes two CNNs with different filter sizes at the first layers and bidirectional-LSTMs. The CNNs can be trained to learn filters to extract time-invariant features from raw single-channel EEG, while the bidirectional-LSTMs can be trained to encode temporal information such as sleep stage transition rules into the model.
  \item We implement a two-step training algorithm that can effectively train our model end-to-end via backpropagation, while preventing the model from suffering class imbalance problem (i.e., learning to classify only the majority of sleep stages) presented in a large sleep dataset.
  \item We show that, without changing the model architecture and the training algorithm, our model could automatically learn features for sleep stage scoring from different raw single-channel EEGs from two datasets, that have different properties (e.g., sampling rate) and scoring standards (AASM and R\&K), without utilizing any hand-engineered features.
\end{itemize}

\IEEEpubidadjcol

\section{DeepSleepNet} \label{sec:deepsleepnet}
\begin{figure}[!t]
\centering
\includegraphics[width=0.4\textwidth,trim={4.9cm 1.4cm 4.75cm 1.2cm},clip]{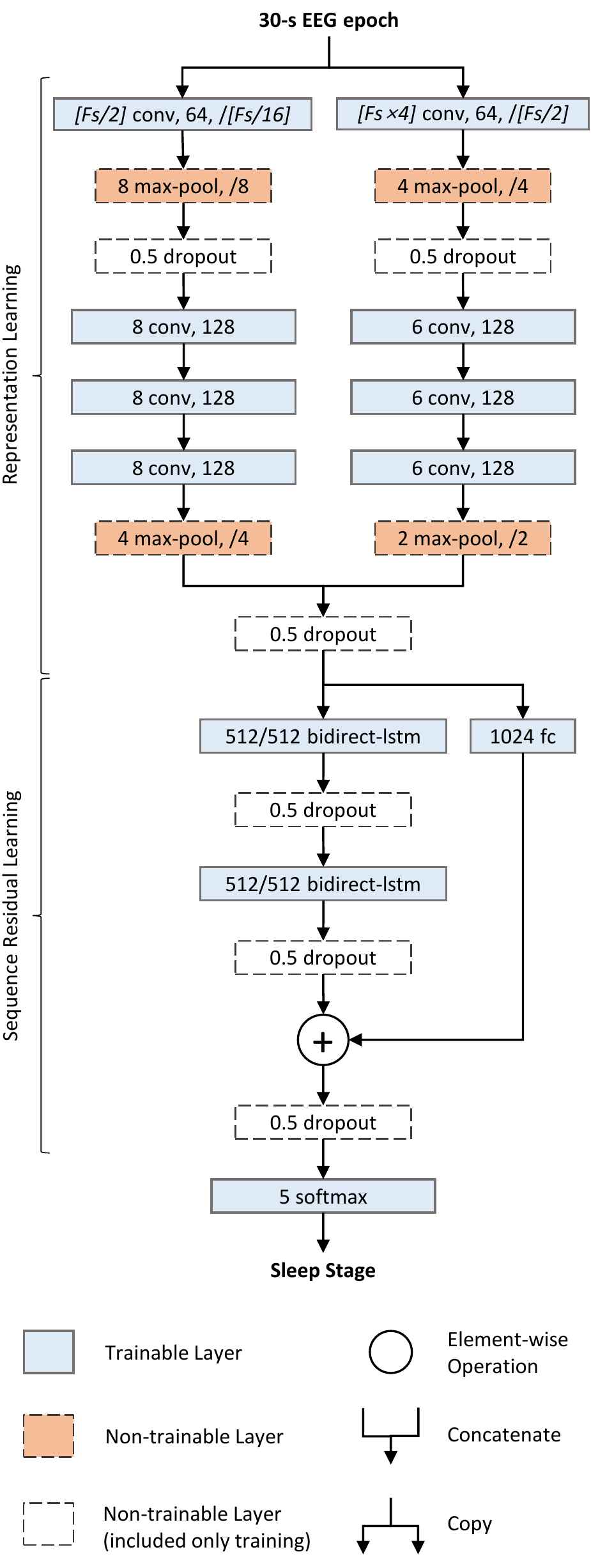}
\caption{An overview architecture of DeepSleepNet consisting of two main parts: representation learning and sequence residual learning. Each trainable layer is a layer containing parameters to be optimized during a training process. The specifications of the first convolutional layers of the two CNNs depends on the sampling rate (Fs) of the EEG data (see Section~\ref{sec:model_spec}).}
\label{fig:deepsleepnet}
\end{figure}

The architecture of DeepSleepNet consists of two main parts as shown in Fig.~\ref{fig:deepsleepnet}.
The first part is representation learning, which can be trained to learn filters to extract time-invariant features from each of raw single-channel EEG epochs. The second part is sequence residual learning, which can be trained to encode the temporal information such as stage transition rules~\cite{iber2007} from a sequence of EEG epochs in the extracted features. This architecture is designed for scoring 30-s EEG epochs following the standard of AASM and R\&K manuals.

\subsection{Representation Learning} \label{sec:representation_learn}
We employ two CNNs with small and large filter sizes at the first layers to extract time-invariant features from raw single-channel 30-s EEG epochs. This architecture is inspired by the way signal processing experts control the trade-off between temporal and frequency precision in their feature extraction algorithms~\cite{cohen2014}. The small filter is better to capture temporal information (i.e., when certain of EEG patterns appear), while the larger filter is better to capture frequency information (i.e., frequency components).

In our model, each CNN consists of four convolutional layers and two max-pooling layers. Each convolutional layer performs three operations sequentially: 1D-convolution with its filters, batch normalization~\cite{ioffe2015}, and applying the rectified linear unit (ReLU) activation (i.e., $relu(x)=max(0,x)$). Each pooling layer downsamples inputs using $max$ operation. The specifications of the filter sizes, the number of filters, stride sizes and pooling sizes can be found in Fig.~\ref{fig:deepsleepnet}. Each \textit{conv} block shows a filter size, the number of filters, and a stride size. Each \textit{max-pool} block shows a pooling size and a stride size. We will explain \textit{dropout} blocks later in Section~\ref{sec:regularization}.

Formally, suppose there are $N$ 30-s EEG epochs $\{\textbf{x}_{1},...,\textbf{x}_{N}\}$ from single-channel EEG. We use the two CNNs to extract the $i$-th feature $\textbf{a}_{i}$ from the $i$-th EEG epoch $\textbf{x}_{i}$ as follows:%
\begin{eqnarray}
\textbf{h}^{s}_{i}&{}={}&CNN_{\theta_{s}}(\textbf{x}_{i}) \label{eqn:cnn_small} \\
\textbf{h}^{l}_{i}&{}={}&CNN_{\theta_{l}}(\textbf{x}_{i}) \label{eqn:cnn_large} \\
\textbf{a}_{i}&{}={}&\textbf{h}^{s}_{i}||\textbf{h}^{l}_{i} \label{eqn:two_cnns}
\end{eqnarray}%
where $CNN(x_{i})$ is a function that transform a 30-s EEG epoch $\textbf{x}_{i}$ into a feature vector $\textbf{h}_{i}$ using a CNN, $\theta_{s}$ and $\theta_{l}$ are parameters of the CNNs with small and large filter sizes in the first layer respectively, and $||$ is a concatenate operation that combines the outputs from two CNNs together. These concatenated or linked features $\{\textbf{a}_{1},...,\textbf{a}_{N}\}$ are then forwarded to the sequence residual learning part.

\subsection{Sequence Residual Learning} \label{sec:sequence_res_learn}
We apply the residual learning framework~\cite{he2015} to design our sequence residual learning part. This part consists of two main components: bidirectional-LSTMs~\cite{schuster1997} and a shortcut connection (see Fig.~\ref{fig:deepsleepnet}).

We employ two layers of bidirectional-LSTMs to learn temporal information such as stage transition rules~\cite{iber2007} which sleep experts use to determine the next possible sleep stages based on the previous stages. For instance, the AASM manual suggests that if a subject is in sleep stage N2, continue to score epochs with low amplitude and mixed frequency EEG activity as N2 even though K complexes or sleep spindles are not present. In this case, the bidirectional-LSTMs can learn to remember that it has seen the stage N2, and continue to score successive epochs as N2 if they still detect the low amplitude and mixed frequency EEG activity. Bidirectional-LSTMs extends the LSTM~\cite{hochreiter1997} by having two LSTMs process forward and backward input sequences independently~\cite{schuster1997}. In other words, the outputs from forward and backward LSTMs are not connected to each other. The model is therefore able to exploit information both from the past and the future. We also use peephole connections~\cite{gers2000,sak2014} in our LSTMs which allow their gating mechanism to inspect their current memory cell before the modification.

We use a shortcut connect to reformulate the computation of this part into a residual function. This enables our model to be able to add temporal information it learns from the previous input sequences into the feature extracted from the CNNs. We also use a fully-connected layer in the shortcut connection to transform the features from the CNNs into a vector that can be added to the output from the LSTMs. This layer performs matrix multiplication with its weight parameters, batch normalization, and applying the ReLU activation sequentially.

Formally, suppose there are $N$ features from the CNNs $\{\textbf{a}_{1}, ..., \textbf{a}_{N}\}$ arranged sequentially and $t=1...N$ denotes the time index of 30-s EEG epochs, our sequence residual learning is defined as follows:%
\begin{eqnarray}
\textbf{h}^{f}_{t},\textbf{c}^{f}_{t}&{}={}&LSTM_{\theta_{f}}(\textbf{h}^{f}_{t-1},\textbf{c}^{f}_{t-1},\textbf{a}_{t}) \label{eqn:lstm_fwd} \\
\textbf{h}^{b}_{t},\textbf{c}^{b}_{t}&{}={}&LSTM_{\theta_{b}}(\textbf{h}^{b}_{t+1},\textbf{c}^{b}_{t+1},\textbf{a}_{t}) \label{eqn:lstm_bwd} \\
\textbf{o}_{t}&{}={}&\textbf{h}^{f}_{t}||\textbf{h}^{b}_{t}+FC_{\theta}(\textbf{a}_{t}) \label{eqn:seq_residual}
\end{eqnarray}%
where $LSTM$ represents a function that processes sequences of features $\textbf{a}_{t}$ using the two-layers LSTM parameterized by $\theta_{f}$ and $\theta_{b}$ for forward and backward directions; $\textbf{h}$ and $\textbf{c}$ are vectors of hidden and cell states of the LSTMs; $\textbf{h}^{f}_{0},\textbf{c}^{f}_{0},\textbf{h}^{b}_{N+1}$ and $\textbf{c}^{b}_{N+1}$ of forward and backward LSTMs are set to zero vectors; $FC$ represents a function that transform features $\textbf{a}_{t}$ into a vector that can be added (element-wise) with the concatenated output vector $\textbf{h}^{f}_{t}||\textbf{h}^{b}_{t}$ from the bidirectional-LSTMs. The specifications of the hidden size of forward and backward LSTMs, and the fully-connected layers can be found in Fig.~\ref{fig:deepsleepnet}. Each \textit{bidirect-lstm} block shows hidden sizes of forward and backward LSTMs. Each \textit{fc} block shows a hidden size.

It should be noted that the hidden and cell states $\textbf{h}^{f}_{t},\textbf{h}^{b}_{t},\textbf{c}^{f}_{t}$ and $\textbf{c}^{b}_{t}$ in \eqref{eqn:lstm_fwd} and \eqref{eqn:lstm_bwd} will be re-initialized to zeros at the beginning of each patient data during the training and testing. This is to make sure that the model uses only temporal information from the current subject data for both training and testing.

\subsection{Model Specification} \label{sec:model_spec}
For the representation learning part, the parameters of the CNN-1 and CNN-2 were selected with the aim to capture temporal and frequency information from the EEG according to the guideline provided by~\cite{cohen2014}. For instance, in Fig.~\ref{fig:deepsleepnet}, the filter size of the \textit{conv1} layers of the CNN-1 was set to Fs/2 (i.e., half of the sampling rate (Fs)), and its stride size was set to Fs/16 to detect when certain of EEG patterns appear. On the other hand, the filter size of the \textit{conv1} layer of the CNN-2 was set to Fs$\times$4 to better capture the frequency components from the EEG. Its stride size was also set to Fs/2, which is higher than the \textit{conv1} layer of the CNN-1, as it is not necessary to perform a fine-grained convolution to extract frequency components. The filter and stride sizes of the subsequent convolutional layers \textit{conv2\_[1-3]} were chosen to be small fix sizes. It is believed that the use of multiple convolutional layers with a small filter size instead of a single convolutional layer with a large filter can reduce the number of parameters and the computational cost, and can still achieve the similar level of model expressiveness~\cite{szegedy2015}.

For the sequence residual learning part, the parameters of the \textit{bidirect-lstm} and \textit{fc} layers were set to
be smaller than the output of the representation learning part, which is 1024 in Fig.~\ref{fig:deepsleepnet}.
This is to restrict our model to select and combine only the important features to prevent overfitting.


\section{Two-Step Training Algorithm} \label{sec:two_steps}
The two-step training algorithm (see Algorithm~\ref{alg:two_steps}) is a technique we develop to effectively train our model end-to-end via backpropagation, while preventing the model from suffering class imbalance problem (i.e., learning to classify only the majority of sleep stages) present in a large sleep dataset. The algorithm first pre-trains the representation learning part of the model and then fine-tunes the whole model using two different learning rates. We use the cross-entropy loss to quantify the agreement between the predicted and the target sleep stages in both of these training steps. The combination of the softmax function (i.e., the last layer in Fig.~\ref{fig:deepsleepnet}) and the cross-entropy loss are used to train our model to output probabilities for mutually exclusive classes.

\subsection{Pre-training} \label{sec:two_steps_pretrain}
The first step is to perform a supervised pre-training on the representation learning part of the model with a class-balance training set so that the model does not overfit to the majority of sleep stages. This can be seen in Algorithm~\ref{alg:two_steps}, lines 1-8. Specifically, the two CNNs are extracted from the model and then stacked with a softmax layer, $softmax$.
It is important to note that this $softmax$ is different from the last layer in the model (see Fig.~\ref{fig:deepsleepnet}).
This stacked softmax layer is only used in this step to pre-train the two CNNs, in which its parameters are discarded at the end of the pre-training. We denote these two CNNs stacked with $softmax$ as $pre\_model$. Then the $pre\_model$ is trained with a class-balance training set using a mini-batch gradient-based optimizer called Adam~\cite{kingma2014} with a learning rate, $lr$. At the end of the pre-training, the softmax layer is discarded. The class-balance training set is obtained from duplicating the minority sleep stages in the original training set such that all sleep stage have the same number of samples (i.e., oversampling).

\subsection{Fine-tuning} \label{sec:two_steps_finetune}
The second step is to perform a supervised fine-tuning on the whole model with a sequential training set. This can be seen in Algorithm~\ref{alg:two_steps}, lines 9-19.  This step is to encode the stage transition rules into the model as well as to perform necessary adjustments on the pre-trained CNNs.
Specifically, the parameters $\theta_{s}$ and $\theta_{l}$ of the two CNNs of $init\_model$ are replaced with the ones from the $pre\_model$, resulting in $model$.
Then the $model$ is trained with the sequence training set using a mini-batch Adam optimizer with two different learning rates, $lr_{1}$ and $lr_{2}$. As the CNNs part has already been pre-trained, we, therefore, use a lower learning rate $lr_{1}$ for the CNNs part and a higher learning rate $lr_{2}$ for the sequence residual learning part, and a softmax layer. We found that when we used the same learning rate to fine-tune the whole network, the pre-trained CNN parameters were excessively adjusted to the sequential data, which were not class-balanced. As a consequence, the model started to overfit to the majority of the sleep stages toward the end of the fine-tuning. Therefore, two different learning rates are used during fine-tuning. Also, we use a heuristic gradient clipping technique to prevent the exploding gradients, which is a well-known problem when training RNNs such as LSTMs~\cite{pascanu2012}. This technique rescales the gradients to smaller values using their global norm whenever they exceed a pre-defined threshold. The sequential training set is obtained by arranging the original training set sequentially according to time across all subjects.

\subsection{Regularization} \label{sec:regularization}
We employed two regularization techniques to help prevent overfitting problems. The first technique was dropout~\cite{srivastava2014,zaremba2014} that randomly sets the input values to 0 (i.e., dropping units along with their connection) with the specified probability during training. Dropout layers with the probability of 0.5 were used throughout the model as shown in Fig.~\ref{fig:deepsleepnet}. 
It is important to note that these dropout layers were used for training only, and were removed from the model during testing to provide deterministic outputs. 

The second technique was L2 weight decay, which adds a penalty term into a loss function to prevent large values of the parameters in the model (i.e., exploding gradients). We only applied the weight decay on the first layers of the two CNNs because of the two main reasons. Firstly, it is pointed out in~\cite{pascanu2012} that L2 weight decay can limit the model capabilities of learning long-term dependencies.
Secondly, we found that, without weight decay, the filters of the first layers of the CNNs overfitted to noises or artifacts in EEG data. This weight decay helped the model learn smoother filters (i.e., containing less high-frequency elements) which resulted in slightly performance gains. The weight decay parameter that defines the degree of penalty, lambda, was set to $10^{-3}$.

\begin{algorithm}[!t]
\caption{Two-step Training}
\label{alg:two_steps}
\begin{algorithmic}[1]
\renewcommand{\algorithmicrequire}{\textbf{Input:}}
\renewcommand{\algorithmicensure}{\textbf{Output:}}
\REQUIRE $init\_model$, $data$
\ENSURE $model$
\\ \textit{Initialization}:
\STATE $init\_CNN_{\theta_{s},\theta_{l}} \gets extract\_cnns(init\_model)$
\STATE $pre\_model \gets stack(init\_CNN_{\theta_{s},\theta_{l}},softmax)$
\STATE $data_{over} \gets oversample(data)$
\\ \textit{Pre-training Step}:
\FOR {$i=1$ \textbf{to} $n\_pretrain\_epochs$}
\FOR {\textbf{each} $batch$ \textbf{in} $shuffle(data_{over})$}
\STATE $pre\_model \gets adam_{lr}(pre\_model,batch)$
\ENDFOR
\ENDFOR
\\ \textit{Fine-tuning Step}:
\STATE $pre\_CNN_{\theta_{s},\theta_{l}} \gets extract\_cnns(pre\_model)$
\STATE $model \gets replace\_cnns(init\_model, pre\_CNN_{\theta_{s},\theta_{l}})$
\FOR {$i=1$ \textbf{to} $n\_finetune\_epochs$}
\FOR {\textbf{each} $subject$ \textbf{in} $data$}
\STATE $model \gets reset\_lstm\_cell\_state(model)$
\STATE $subject\_data_{seq} \gets arrange\_sequence(subject)$
\FOR {\textbf{each} $batch$ \textbf{in} $subject\_data_{seq}$}
\STATE $model \gets adam_{lr_{1},lr_{2}}(model,batch)$
\ENDFOR
\ENDFOR
\ENDFOR
\RETURN $model$
\end{algorithmic}
\end{algorithm}

\section{Results} \label{sec:results}

\subsection{Data} \label{sec:data}
We evaluated our model using different EEG channels from two public datasets: Montreal Archive of Sleep Studies (MASS)~\cite{oreilly2014} and Sleep-EDF~\cite{goldberger2000,kemp2000}.

\textbf{MASS.} In MASS cohort 1, there were five subsets of recordings, SS1-SS5, which were organized according to their research and acquisition protocols. We used data from SS3, which contained PSG recordings from 62 healthy subjects (age 42.5$\pm$18.9). Each recording contained 20 scalp-EEG, 2 EOG (left and right), 3 EMG and 1 ECG channels. The EEG electrodes were positioned according to the international 10-20 system, and the EOG electrodes were positioned diagonally on the outer edges of the eyes. EEG and EOG recordings were pre-processed with a notch filter of 60 Hz, and band-pass filters of 0.30-100 Hz (EEG) and 0.10-100 Hz (EOG). All EEG and EOG recordings had the same sampling rate of 256 Hz. These recordings were manually classified into one of the five sleep stages (W, N1, N2, N3 and REM) by a sleep expert according to the AASM standard~\cite{iber2007}. There were also movement artifacts at the beginning and the end of each subject's recordings that were labeled as UNKNOWN. We evaluated our model using the F4-EOG (Left) channel, which was obtained via montage reformatting~\cite{lagerlund2000} without any further pre-processing. 

\textbf{Sleep-EDF.} There were two sets of subjects from two studies: age effect in healthy subjects (SC) and Temazepam effects on sleep (ST). We used 20 subjects (age 28.7$\pm$2.9) from SC. Each PSG recording contained 2 scalp-EEG signals from Fpz-Cz and Pz-Cz channels, 1 EOG (horizontal), 1 EMG, and 1 oro-nasal respiration signal. All EEG and EOG had the same sampling rate of 100 Hz. These recordings were manually classified into one of the eight classes (W, N1, N2, N3, N4, REM, MOVEMENT, UNKNOWN) by sleep experts according to the R\&K standard~\cite{AllanHobson1969}. We evaluated our model using the Fpz-Cz and Pz-Cz channels without any further pre-processing. We also merged the N3 and N4 stages into a single stage N3 to use the same AASM standard as the MASS dataset. There were long periods of awake or stage W at the start and the end of each recording, in which a subject was not sleeping. We only included 30 minutes of such periods just before and after the sleep periods, as we were interested in sleep periods.

We excluded MOVEMENT and UNKNOWN (which were at the start or the end of the each recording) stages, as they did not belong to the five sleep stages~\cite{iber2007}. Table~\ref{tab:dataset} summarizes the number of 30-s epochs for each sleep stage from these two datasets.

\begin{table}[!t]
\renewcommand{\arraystretch}{1.3}
\caption{Number of 30-s epochs for each sleep stage from two datasets}
\label{tab:dataset}
\centering
\begin{tabular}{|l|cccccc|}
\hline
Dataset   & W    & N1   & N2    & N3 (N4) & REM   & Total \\ \hline
MASS      & 6227 & 4724 & 29534 & 7651    & 10464 & 58600 \\
Sleep-EDF & 7927 & 2804 & 17799 & 5703    & 7717  & 41950 \\ \hline
\end{tabular}
\end{table}

\subsection{Experimental Design} \label{sec:experimental_design}
We evaluated our model using a $k$-fold cross-validation scheme, where $k$ was set to 31 and 20 for the MASS and Sleep-EDF datasets respectively. Specifically, in each fold, we used recordings from $N_{s} - (N_{s}/k)$ to train the model, and from the remaining $N_{s}/k$ subjects to test the trained model, where $N_{s}$ is the number of subjects in the dataset. This process was repeated $k$ times so that all of the recordings were tested.
Then we combined the predicted sleep stages from all folds and computed the performance metrics, which will be discussed in Section~\ref{sec:performance_metric}.

\subsection{Performance Metrics} \label{sec:performance_metric}
We evaluated the performance of our model using per-class precision (PR), per-class recall (RE), per-class F1-score (F1), macro-averaging F1-score (MF1), overall accuracy (ACC), and Cohen's Kappa coefficient ($\kappa$)~\cite{cohen1960,sokolova2009}. The per-class metrics are computed by considering a single class as a positive class, and all other classes combined as a negative class. The MF1 and ACC are calculated as follows:
\begin{eqnarray}
\text{ACC}&{}={}&\dfrac{\sum_{c=1}^{C}\text{TP}_{c}}{N} \label{eqn:acc} \\
\text{MF1}&{}={}&\dfrac{\sum_{c=1}^{C}\text{F1}_{c}}{C} \label{eqn:mf1}
\end{eqnarray}
where TP$_{c}$ is the true positives of class $c$, F1$_{c}$ is per-class F1-score of class $c$, $C$ is the number of sleep stages, and $N$ is the total number of test epochs.

\subsection{Training Parameters} \label{sec:training_params}
The representation learning part was pre-trained using the oversampled training set with the mini-batch size of 100. The Adam optimizer's parameters $lr$, $beta1$, and $beta2$ were set to $10^{-4}$, 0.9 and 0.999 respectively. Then the whole model was fine-tuned using the sequential training set. Specifically, we equally split the sequences of 30-s EEG epochs from each subject data into 10 sub-sequences (i.e., batch size was 10). Then we fed 25 epochs (i.e., sequence length was 25) from each sub-sequence yielding 250 epochs per one step training. The Adam optimizer's parameters were similar to the pre-training step except that the learning rate of each part of the model, $lr1$ and $lr2$, were set to $10^{-6}$ and $10^{-4}$ respectively. The threshold of the gradient clipping was set to 10. The numbers of epochs for the pre-training and the fine-tuning steps were set to 100 and 200 respectively. There was no early stopping as there was no validation set in our evaluation scheme.


For the batch normalization in \textit{conv} and \textit{fc} blocks, the $\epsilon$ constant of 10$^{-5}$ was added to the mini-batch variance for numerical stability. The mean and variance of the training set, which were used as fixed parameters during testing, were estimated by computing the moving average of with a decay rate of 0.999 from the sampling mean and variance of each mini-batch.

\subsection{Implementation} \label{sec:implementation}
We implemented our model using TensorLayer (\url{https://github.com/zsdonghao/tensorlayer}), which is a deep learning library extended from Google Tensorflow~\cite{abadi2015}. This library allows us to deploy numerical computation such as the training and validation tasks to multiple CPUs and GPUs. We ran the k-fold cross-validation using the eTRIKS Analytical Environment (eAE) (\url{https://eae.doc.ic.ac.uk/}), which provides a cluster of high-performance computing nodes. Each node was equipped with an NVIDIA GeForce GTX 980. The training time for each validation fold was approximately 3 hours on each node. The testing or prediction time for each batch of 25 EEG epochs (according to the sequence length specified during training in Section~\ref{sec:training_params}) was approximately 50 milliseconds on each node.

\subsection{Initial Experiments} \label{sec:init_exp}
We initially conducted experiments with the first fold of the 31-fold cross-validation with the MASS dataset. This was to design the architecture and the parameters for DeepSleepNet.
For model architecture, we tried several configurations such as increasing/decreasing convolutional layers, changing the number of filters and the stride sizes, and changing the number of hidden sizes in the bidirectional-LSTMs and the fully-connected layer. The architecture in Fig.~\ref{fig:deepsleepnet} gave us the best performance.
For regularization parameters, we tried several values for the weight decay parameters ranging from $10^{-1}$ to $10^{-5}$. The value of $10^{-3}$ gave us the best performance.
For training parameters, we tried several values of learning rates ranging from $10^{-3}$ to $10^{-8}$. We also experimented with the mini-batch size (from 50 to 200) during the pre-training, the batch size (from 5 to 40) and sequence length (from 5 to 40) during fine-tuning. Other parameters such as $beta1$, $beta2$ and the threshold of the gradient clipping were chosen from the default values reported in the literature. The training parameters mentioned in Section~\ref{sec:training_params} gave us the best performance. With these settings, the pre-training and fine-tuning steps started to converge after 100 and 200 epochs respectively.

\subsection{Sleep Stage Scoring Performance} \label{sec:performance}
Table~\ref{tab:cm_deepsleepnet_mass_f4} and \ref{tab:cm_deepsleepnet_sleepedf_fpz} show confusion matrices obtained from the 31-fold and the 20-fold cross-validation on the F4-EOG (Left) and the Fpz-Cz channels from the MASS and Sleep-EDF datasets respectively. We did not include the confusion matrix obtained from the Pz-Oz channel from the Sleep-EDF dataset as the Fpz-Cz channel gave a better performance.
Each row and column represent the number of 30-s EEG epochs of each sleep stage classified by the sleep expert and our model respectively. The numbers in bold indicate the number of epochs that were correctly classified by our model. The last three columns in each row indicate per-class performance metrics computed from the confusion matrix.

It can be seen that the poorest performance was noted for the stage N1, with the F1 less than 60, while the F1 for other stages were significantly better, with the range between 81.5 and 90.3.
Most of the misclassified stages were between N2 and N3. It can also be seen that the confusion matrix is almost symmetric via the diagonal line (except the pair of N2-N3). This indicates that the misclassifications were less likely to be due to the imbalance-class problem.

\begin{figure*}[!t]
\centering
\includegraphics[width=\textwidth]{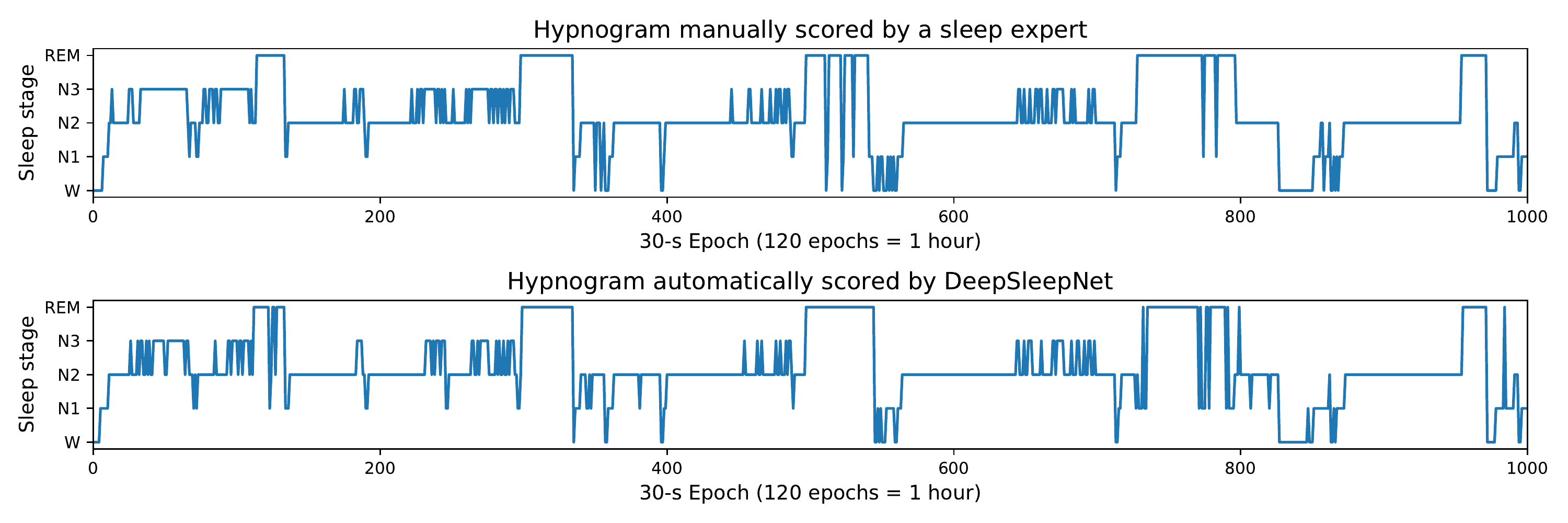}
\caption{Examples of the hypnogram manually scored by a sleep expert (top) and the hypnogram automatically scored by DeepSleepNet (bottom) for Subject-1 from the MASS dataset.}
\label{fig:hypnogram}
\end{figure*}

Fig.~\ref{fig:hypnogram} demonstrate examples of hypnograms that were manually scored by a sleep expert, and automatically scored by our DeepSleepNet for Subject-1 from the MASS dataset.

\subsection{Comparison with State-of-the-Art Approaches}
Table~\ref{tab:comparison} shows a comparison between our method and other sleep stage scoring methods across ACC, MF1, $\kappa$ and F1. These methods include the ones that utilize hand-engineered features~\cite{hsu2013,sharma2017,hassan2017,tsinalis2016}, CNNs only~\cite{tsinalis2016cnn} or LSTMs only (which is our previous work)~\cite{dong2016lstm}.
The other methods' metrics were computed using the confusion matrices reported in their papers. We classified the methods into two groups: \textit{non-independent} and \textit{independent} training and test sets. The non-independent ones were the methods that included parts of the test subjects' epochs in the training data, while the independent ones were the methods that excluded all epochs of the test subjects from the training data. We believe that the practical evaluation scheme should not include any epochs from the test subjects. Also, it has been shown that the non-independent scheme resulted in an improvement of the performance~\cite{tsinalis2016}. Thus we did not compare the performance of our method with the non-independent group. The numbers in bold indicate the highest performance metrics of all methods in each dataset of each group.

Among the methods in the independent group, it can be seen that our method achieved a similar performance compared to the state-of-the-art methods that used the same EEG channel and dataset. It should also be emphasized that our model achieved the similar performance without sacrificing the performance on the stage N1, which is the most difficult sleep stage to classify. This indicates that our method was not biased by favoring the majority of the sleep stages than the minority ones. According to the $\kappa$ coefficient, it showed that the agreement between the sleep experts and our model were substantial (between 0.61 and 0.80)~\cite{hassan2017}. It should also be noted that our model performed better when applied on the Fpz-Cz channel compared to the Pz-Oz, which is similar to~\cite{tsinalis2016}.

\begin{table}[!t]
\renewcommand{\arraystretch}{1.3}
\caption{Confusion matrix obtained from 31-fold cross-validation on F4-EOG (Left) channel from the MASS dataset}
\label{tab:cm_deepsleepnet_mass_f4}
\centering
\begin{tabular}{|c|ccccc|ccc|}
\hline
 & \multicolumn{5}{c|}{Predicted} & \multicolumn{3}{c|}{Per-class Metrics} \\
 & W & N1 & N2 & N3 & REM & PR & RE & F1 \\ \hline
W & \textbf{5433} & 572 & 107 & 13 & 102 & 87.3 & 87.2 & 87.3 \\
N1 & 452 & \textbf{2802} & 827 & 4 & 639 & 60.4 & 59.3 & 59.8 \\
N2 & 185 & 906 & \textbf{26786} & 1158 & 499 & 89.9 & 90.7 & 90.3 \\
N3 & 18 & 4 & 1552 & \textbf{6077} & 0 & 83.8 & 79.4 & 81.5 \\
REM & 132 & 356 & 533 & 1 & \textbf{9442} & 88.4 & 90.2 & 89.3 \\ \hline
\end{tabular}
\end{table}

\begin{table}[!t]
\renewcommand{\arraystretch}{1.3}
\caption{Confusion matrix obtained from 20-fold cross-validation on Fpz-Cz channel from the Sleep-EDF dataset}
\label{tab:cm_deepsleepnet_sleepedf_fpz}
\centering
\begin{tabular}{|c|ccccc|ccc|}
\hline
\textbf{} & \multicolumn{5}{c|}{\textbf{Predicted}}                                        & \multicolumn{3}{c|}{Per-class Metrics} \\
          & W             & N1            & N2             & N3            & REM           & PR          & RE          & F1         \\ \hline
W         & \textbf{6614} & 745           & 181            & 81            & 306           & 86.0        & 83.4        & 84.7       \\
N1        & 295           & \textbf{1406} & 631            & 30            & 442           & 43.5        & 50.1        & 46.6       \\
N2        & 391           & 618           & \textbf{14542} & 1473          & 775           & 90.5        & 81.7        & 85.9       \\
N3        & 29            & 9             & 291            & \textbf{5370} & 4             & 77.1        & 94.2        & 84.8       \\
REM       & 360           & 457           & 419            & 7             & \textbf{6474} & 80.9        & 83.9        & 82.4       \\ \hline
\end{tabular}
\end{table}

\begin{table*}[!t]
\renewcommand{\arraystretch}{1.3}
\caption{Comparison between DeepSleepNet and other sleep stage scoring methods that utilizes hand-engineering features across overall accuracy (ACC), macro-F1 score (MF1), Cohen's Kappa ($\kappa$), and Per-class F1-Score (F1)}
\label{tab:comparison}
\centering
\begin{tabular}{|c|c|c|c|ccc|ccccc|}
\hline
\multirow{2}{*}{Methods} & \multirow{2}{*}{Dataset} & \multirow{2}{*}{EEG Channel} & \multirow{2}{*}{Test Epochs} & \multicolumn{3}{c|}{Overall Metrics} & \multicolumn{5}{c|}{Per-class F1-Score (F1)} \\
 &  &  &  & ACC & MF1 & $\kappa$ & W & N1 & N2 & N3 & REM \\ \hline
\multicolumn{12}{|l|}{\textit{Non-independent Training and Test Sets}} \\ \hline
Ref.~\cite{hsu2013} & Sleep-EDF & Fpz-Cz & 960 & 90.3 & 76.5 & - & 77.3 & 46.5 & \textbf{94.9} & 72.2 & \textbf{91.8} \\
Ref.~\cite{sharma2017} & Sleep-EDF & Pz-Oz & 15136 & \textbf{91.3} & 77 & \textbf{0.86} & \textbf{97.8} & 30.4 & 89 & \textbf{85.5} & 82.5 \\
Ref.~\cite{hassan2017} & Sleep-EDF & Pz-Oz & 7596 & 90.8 & \textbf{80} & 0.85 & 96.9 & \textbf{49.1} & 89 & 84.2 & 81.2 \\ \hline
\multicolumn{12}{|l|}{\textit{Independent Training and Test Sets}} \\ \hline
Ref.~\cite{tsinalis2016} & Sleep-EDF & Fpz-Cz & 37022 & 78.9 & 73.7 & - & 71.6 & \textbf{47.0} & 84.6 & 84.0 & 81.4 \\
Ref.~\cite{tsinalis2016cnn} & Sleep-EDF & Fpz-Cz & 37022 & 74.8 & 69.8 & - & 65.4 & 43.7 & 80.6 & \textbf{84.9} & 74.5 \\
DeepSleepNet & Sleep-EDF & Fpz-Cz & 41950 & \textbf{82.0} & \textbf{76.9} & \textbf{0.76} & 84.7 & 46.6 & \textbf{85.9} & 84.8 & \textbf{82.4} \\
DeepSleepNet & Sleep-EDF & Pz-Oz & 41950 & 79.8 & 73.1 & 0.72 & \textbf{88.1} & 37 & 82.7 & 77.3 & 80.3 \\
\rule{0pt}{4ex} Ref.~\cite{dong2016lstm} & MASS & F4-EOG (Left) & 59066 & 85.9 & 80.5 & - & 84.6 & 56.3 & \textbf{90.7} & \textbf{84.8} & 86.1 \\
DeepSleepNet & MASS & F4-EOG (Left) & 58600 & \textbf{86.2} & \textbf{81.7} & \textbf{0.80} & \textbf{87.3} & \textbf{59.8} & 90.3 & 81.5 & \textbf{89.3} \\ \hline
\end{tabular}
\end{table*}

\subsection{Sequence Residual Learning} \label{sec:seq_res_learn}
We performed additional experiments to verify the important of the sequence residual learning part with the MASS dataset.
Table~\ref{tab:cm_deepsleepnet_mass_f4_nonseq} shows a confusion matrix obtained from 31-fold cross-validation on the F4-EOG (Left) channel using DeepSleepNet without the sequence residual learning part (i.e., using the $pre\_model$ in Algorithm~\ref{alg:two_steps}). It can be seen that the F1 of all sleep stages, except the stage N3, were lower than the ones in Table~\ref{tab:cm_deepsleepnet_mass_f4}. This was because of an increase in the misclassifications between the pairs of N1-N2, N2-N3 and N1-REM. This may well be due to the effects of oversampling the training set to have balanced-class samples. As a consequence the model tended to predict more of stages N1 and N3. These results indicated that the process to stack the pre-trained representation learning part with the sequence residual learning part, and then fine-tune the both parts with sequential training set helped improve the classification performance.

\subsection{Model Analysis} \label{sec:model_analysis}
To better understanding how our model classified a sequence of 30-s EEG epochs, we analyzed and compared: 1) the learned filters at the first convolutional layers of the two CNNs in the representation learning part; and 2) the memory cells inside the bidirectional-LSTMs in the sequence residual learning part. This analysis was carried out with the MASS dataset across 31 cross-validation folds.

\begin{figure*}[!t]
\centering
\subfloat[]{\includegraphics[width=0.48\textwidth,trim={0cm 2.4cm 0cm 2.3cm},clip]{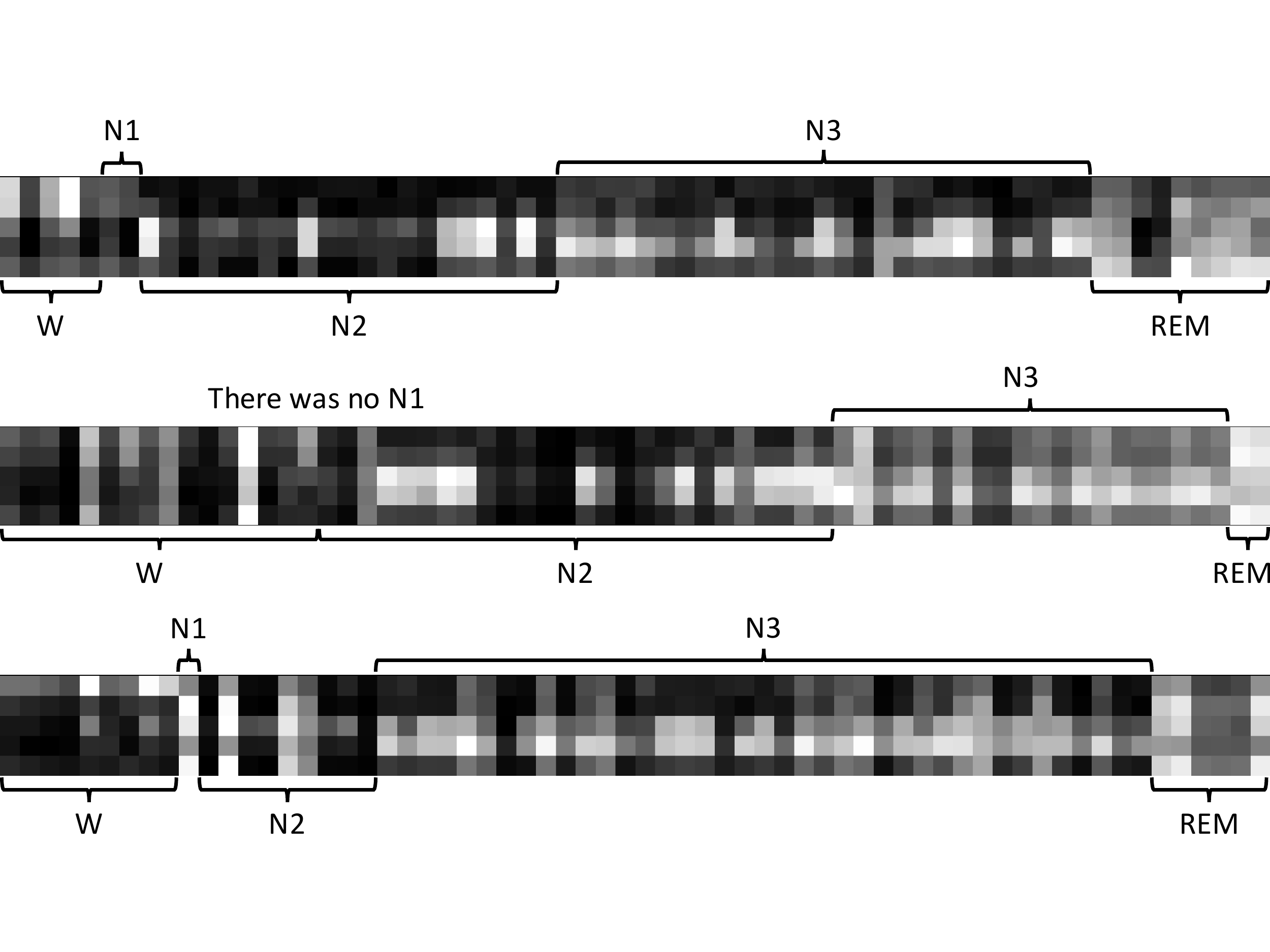}%
\label{fig:small_active_filters}}
\hfil
\subfloat[]{\includegraphics[width=0.48\textwidth,trim={0cm 2.4cm 0cm 2.3cm},clip]{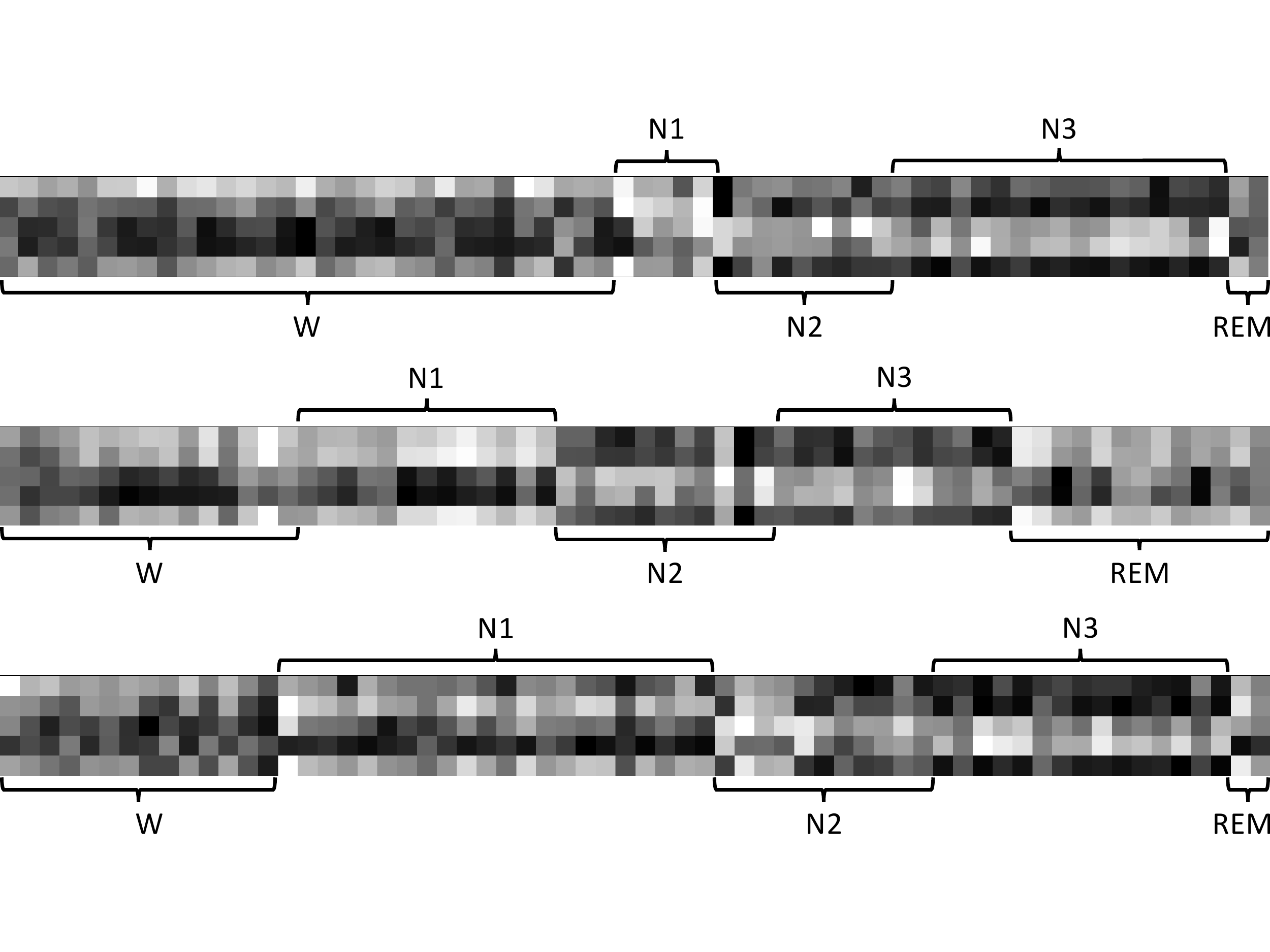}%
\label{fig:large_active_filters}}
\caption{Examples of the filter activations from the first convolutional layers of the two CNNs obtained by feeding our model with data from 3 subjects. The filter activations from the small filters are on the left (a), and the larger filters are on the right (b). Each image has 5 rows and 64 columns, corresponding to 5 sleep stages and 64 filters respectively. Each pixel represents the scaled value of $u_{c,k}$ from \eqref{eqn:filter_act}, where 1 (i.e., active) is white and 0 (i.e., inactive) is black. Each row corresponds to the $64$-dimension vector (i.e., $k$ is 64) for each sleep stage $c$. The first row is from stage W and the last row is from stage REM. Each image also has labels indicating which filters are mostly active for which sleep stages.}
\label{fig:active_filters}
\end{figure*}

Firstly, we analyzed how our model utilized the learned filter at the first convolutional layers of the two CNNs to classify different sleep stages. Specifically, we determined which filters were mostly active for each sleep stage by computing the average of the sum of the activations of all filters across samples of each sleep stage. Formally, suppose there were $N$ 30-s EEG epochs from each validation fold $\{\textbf{x}_{1}, ..., \textbf{x}_{N}\}$. We fed these epochs to our model to obtain activations $\textbf{z}$ from the first convolutional layer of each CNN: $\{\textbf{z}_{1},..., \textbf{z}_{N}\}$, where $\textbf{z}_{i} \in \mathbb{R}^{p \times q}$, and $p$ and $q$ are the activation output size and the number of filters of the first convolutional layer. The average of the sum of the activations of the filter $k$ for the sleep stage $c$ is computed as follows:%
\begin{eqnarray}
u_{c,k}&{}={}&\dfrac{\sum_{i=1}^{N_{y_{pred}=c}} \sum_{j=1}^{q} z_{i,j,k}}{N_{y_{pred}=c}}  \label{eqn:filter_act}
\end{eqnarray}%
where $u_{c,k}$ is the average of the sum of the activation of the filter $k$ for sleep stage $c$, $z_{i,j,k}$ is the $j$-th index of the activation vector $\textbf{z}_{i}$ of the filter $k$, and $N_{y_{pred}=c}$ is the number of EEG epochs that our model predicted as stage $c$. After we computed the $u_{c,k}$ of all filters for sleep stage $c$, we rescaled them into a range of 0 and 1. We denote this scaled $k$-dimension vector $\textbf{u}_{c}$ as \textit{filter activations} for stage $c$. This process was repeated for all sleep stages. Once we got the filter activations from all sleep stages, we stacked them together, and rearranged the order of the filters such that the filters that were mostly active for each sleep stage were grouped together. Fig.~\ref{fig:active_filters} illustrates an example of the filter activations from the small (a) and large (b) filters obtained by feeding our model with data from 3 subjects. Each image has 5 rows and 64 columns, corresponding to 5 sleep stages and 64 filters respectively. Each pixel represents the value of $u_{c,k}$ from \eqref{eqn:filter_act} scaled into a range of 0 and 1, where 1 (i.e., active) is white and 0 (i.e., inactive) is black. Each row corresponds to the $64$-dimension vector (i.e., $k$ is 64) for each sleep stage $c$. The first row is from stage W and the last row is from stage REM. Each image also has labels indicating which filters are mostly active for which sleep stages. We found that there were two types of filters: ones that were mostly active for each sleep stage, and the other ones that were mostly active for multiple sleep stages. For instance, some of the small and large filters were mostly active for both stage N2 and N3. After we had analyzed all of the filter activations from different cross-validation folds, we found that the number of active filters for different sleep stages varied across subjects, and most of the small filters were mostly active for stage N2 and N3. We also found that, for a few subjects, no small filter was active for stage N1. This might well be because there were only a few stage N1 in the dataset.

\begin{table}[!t]
\renewcommand{\arraystretch}{1.3}
\caption{Confusion matrix obtained from 31-fold cross-validation on the F4-EOG (Left) channel from the MASS dataset using DeepSleepNet without Sequence Residual Learning}
\label{tab:cm_deepsleepnet_mass_f4_nonseq}
\centering
\begin{tabular}{|c|ccccc|ccc|}
\hline
 & \multicolumn{5}{c|}{Predicted} & \multicolumn{3}{c|}{Per-class Metrics} \\
 & W & N1 & N2 & N3 & REM & PR & RE & F1 \\ \hline
W & \textbf{5215} & 709 & 94 & 19 & 190 & 84.5 & 83.7 & 84.1 \\
N1 & 468 & \textbf{2582} & 747 & 11 & 916 & 40.8 & 54.7 & 46.8 \\
N2 & 241 & 1846 & \textbf{24140} & 2435 & 872 & 93.4 & 81.7 & 87.2 \\
N3 & 19 & 3 & 472 & \textbf{7156} & 1 & 74.3 & 93.5 & 82.8 \\
REM & 227 & 1181 & 383 & 5 & \textbf{8668} & 81.4 & 82.8 & 82.1 \\ \hline
\end{tabular}
\end{table}

\begin{figure}[!t]
\centering
\includegraphics[width=0.48\textwidth]{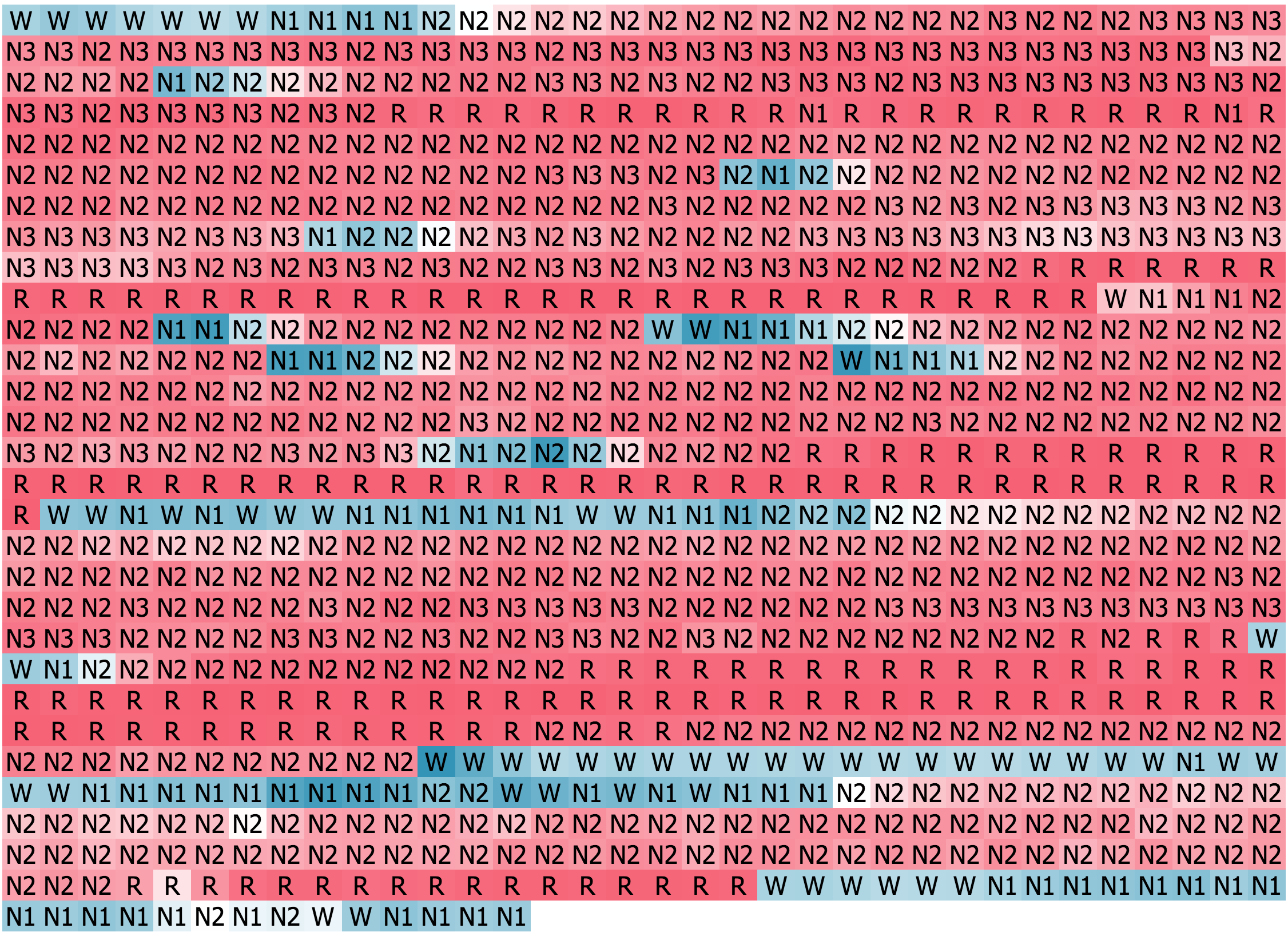}
\caption{An example of the LSTM cell that is active at the beginning of wakefulness (i.e., stage W) or the sleep onset (i.e., stage N1). The sequences of sleep stages are the predictions from DeepSleepNet on one subject data, arranged through time from left-to-right and top-to-bottom. The background color of each stage corresponds to $tanh(c)$, where +1 is blue and -1 is red.}
\label{fig:sleep_onset_cell}
\end{figure}

Secondly, we analyzed how our model utilized the bidirectional-LSTMs to learn the temporal information from a sequence of EEG epochs. Specifically, we investigated how the bidirectional-LSTMs managed their memory cells (i.e., $c$ in \eqref{eqn:lstm_fwd} and \eqref{eqn:lstm_bwd}) using the visualization technique from~\cite{karpathy2015}.  We found several memory cells of the forward LSTMs that were interpretable. For instance, several cells were keeping track of the wakefulness or the sleep onset, which reset their values to positive numbers (i.e., active) when a subject was in the stage W or N1 respectively. The cell values then decreased to negative values (i.e., becoming inactive) during stages N2, N3 and REM (or R in short). Fig.~\ref{fig:sleep_onset_cell} illustrates the changes of this cell value according to a sequence of sleep stages predicted by our model. The sleep stages are arranged according through time from left-to-right, and top-to-bottom. Each sleep stage color corresponds to $tanh(c)$, where +1 (i.e., active) is blue and -1 is red (i.e., inactive). There were also other interpretable cells such as the ones that started with a high value at the beginning of each subject data and then slowly decreased with each sleep stage until the end of the subject data, or the ones that turned on when they found a continuous sequence of stages N3 and REM. The existence of these cells showed that the LSTMs inside the sequence residual learning part learned to keep track of the current status of each subject, which is important to correctly identify the next sleep stages according to the stage transition rules~\cite{iber2007}.

\section{Discussion}
We propose the DeepSleepNet model that utilizes CNNs and bidirectional-LSTMs to automatically learn features for sleep stage scoring from raw single-channel EEGs without using any hand-engineered features. The results showed that, without changing the model architecture and the training algorithm, the model could be applied on different EEG channels (F4-EOG (Left), Fpz-Cz and Pz-Oz). It achieved similar overall accuracy and macro F1-score compared to the state-of-the-art hand-engineering methods on both the MASS and Sleep-EDF datasets, which have different properties such as sampling rate and scoring standards (AASM and R\&K). The results also showed that the temporal information learned from the sequence residual learning part helped improve the classification performance. These demonstrated that our model could automatically learn features for sleep stage scoring from different raw single-channel EEGs.

There are two main reasons that we evaluated our model with the F4-EOG (Left) channel from the MASS dataset, which is different from most of the existing methods reported in the literature that rely on the electrodes at the central lobe such as Cz, C4 and C3. The first reason is to compare the scoring performance with our previous hand-engineering approach. The second reason is that it is much easier and more comfortable to collect data either at sleep clinics or from home environment compared to the existing methods.
This is because both of the electrodes do not have problems of reading the electrical activity from the hairy scalp. Even though the F4-EOG (Left) channel does not have information from the central and occipital lobes as recommended in the AASM manual~\cite{iber2007}, our results showed that our model was still able to achieve a similar performance compared to the state-of-the-art methods.

Based on the results of our simple model analysis, we found that our model learned several interesting features that were consistent with the AASM manual (which is the same manual the experts followed to score the MASS dataset). In the representation learning part, some of the learned filters at the first convolutional layers of the two CNNs were mostly active for stage N2-N3 and W-N1-REM (see Fig~\ref{fig:active_filters}). This implies that our model recognized some patterns that are similar among such stages. Our model might learn the filters to detect sleep spindles that can appear in both N2 and N3 stages, and to detect different features of the eye movements from the EOG (Left) that can be used to distinguish among W, N1 and REM stages. Also, in the sequence residual learning part, we found some interpretable memory cells in the bidirectional-LSTMs such as the cells that were keeping track of the wakefulness or the sleep onset, the cells that increased or decreased its value over time, and the cells that detected a train of stage N3 and REM. Our model utilized a combination of these cells to understand the current status of each subject, and to formulate transition rules. For instance, our model might remember that the subject was now awake or in the stage W. The next possible stage was very likely to be either stage W or N1. It should be emphasized that our model can learn these features from raw single-channel EEG without utilizing any hand-engineered features. Moreover, we observed that the features that our model learned were consistent across different folds. Therefore, we believe that DeepSleepNet is a better approach to implement automatic sleep stage scoring system compared to the hand-engineering ones that require prior knowledge to design feature extraction algorithms.

Even though our results are encouraging, our model is still subject to several limitations. Firstly, our model requires being trained with a sufficient amount of sleep dataset. This is due to the nature of the deep learning techniques that require a significant amount of training data to learn useful representations from the data.
We performed additional experiments with the MASS dataset to estimate the number of epochs required to train our DeepSleepNet. We tried different numbers of folds (i.e., $k$) for the $k$-fold cross-validation from 2 to 31. We found that the scoring performance started to drop when $k$ was less than 12, or when the number of training epochs was approximately less than 54000.
Secondly, as our model learns features from the training data, it might not perform well when the trained model is applied to the data that have properties different from the training data such as data from different EEG channels.
The model might have to be re-trained or fine-tuned before it can be applied to the data with different properties.
Lastly, as our model utilizes bidirectional-LSTMs, the model has to wait until it has collected enough 30-s EEG epochs (depending on the sequence length of the EEG epochs used during the training process) before it can score these epochs. For instance, when the sequence length is set to 25, the model has to wait for 30$\times$25 seconds (or 12.5 minutes) before it can identify sleep stages for these 25 EEG epochs.

\section{Conclusion and Future Work}
We propose a deep learning model, named DeepSleepNet, for automatic sleep stage scoring based on raw single-channel EEG without utilizing any hand-engineered features. Our model utilizes CNNs to extract time-invariant features, and bidirectional-LSTMs to learn stage transition rules among sleep stages from EEG epochs. We also implement the two-step training algorithm that pre-trains our model with the oversampled dataset to alleviate class-imbalance problems, and fine-tunes the model with the sequences of EEG epochs to encode the temporal information into the model.
Our results showed that, without changing the model architecture and the training algorithm, our model was able to automatically learn features for sleep stage scoring from different raw single-channel EEGs from two datasets that have different properties and scoring standards.
Our model analysis results also demonstrated that our model learned several features that are consistent with the AASM manual. As our model automatically learn features from raw EEG, we believe that DeepSleepNet is a better approach to realize a remote sleep monitoring compared to the hand-engineering ones.

In the future, we plan to improve our DeepSleepNet to be able to apply on the single-channel EEG such as the F4-EOG (Left) and Fp2-EOG (Left) collected from wearable devices.

\section*{Acknowledgment}
The authors would like to thank Douglas McIlwraith and Axel Oehmichen from Imperial College London who reviewed the contents of the paper, and provided support for running the $k$-fold cross-validation.

\ifCLASSOPTIONcaptionsoff
  \newpage
\fi



\bibliographystyle{IEEEtran}
\bibliography{IEEEabrv,main}
%

%

\newpage

\begin{IEEEbiography}[{\includegraphics[width=1in,height=1.25in,clip,keepaspectratio]{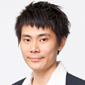}}]{Akara Supratak}
is currently a Ph.D. candidate at Data Science Institute, Department of Computing, Imperial College London. His research is in the area of bio-medical engineering, software engineering and deep learning.
\end{IEEEbiography}

\begin{IEEEbiography}[{\includegraphics[width=1in,height=1.25in,clip,keepaspectratio]{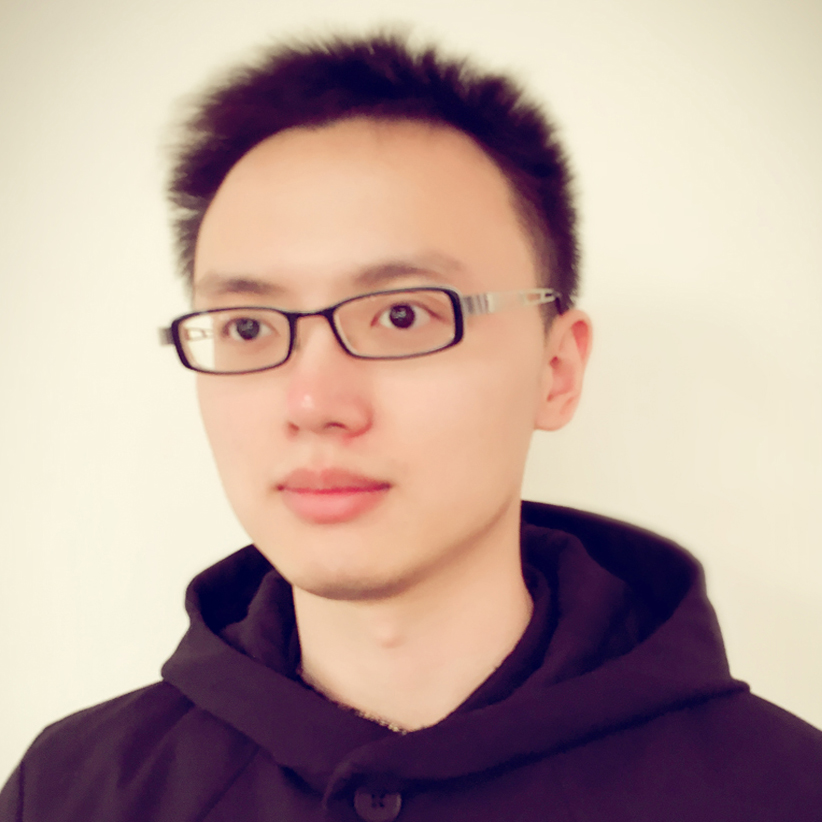}}]{Hao Dong}
is currently a Ph.D. candidate at Data Science Institute, Department of Computing, Imperial College London. His research is in the areas of deep learning and biomedical engineering, especially with EEG data.
\end{IEEEbiography}

\begin{IEEEbiography}[{\includegraphics[width=1in,height=1.25in,clip,keepaspectratio]{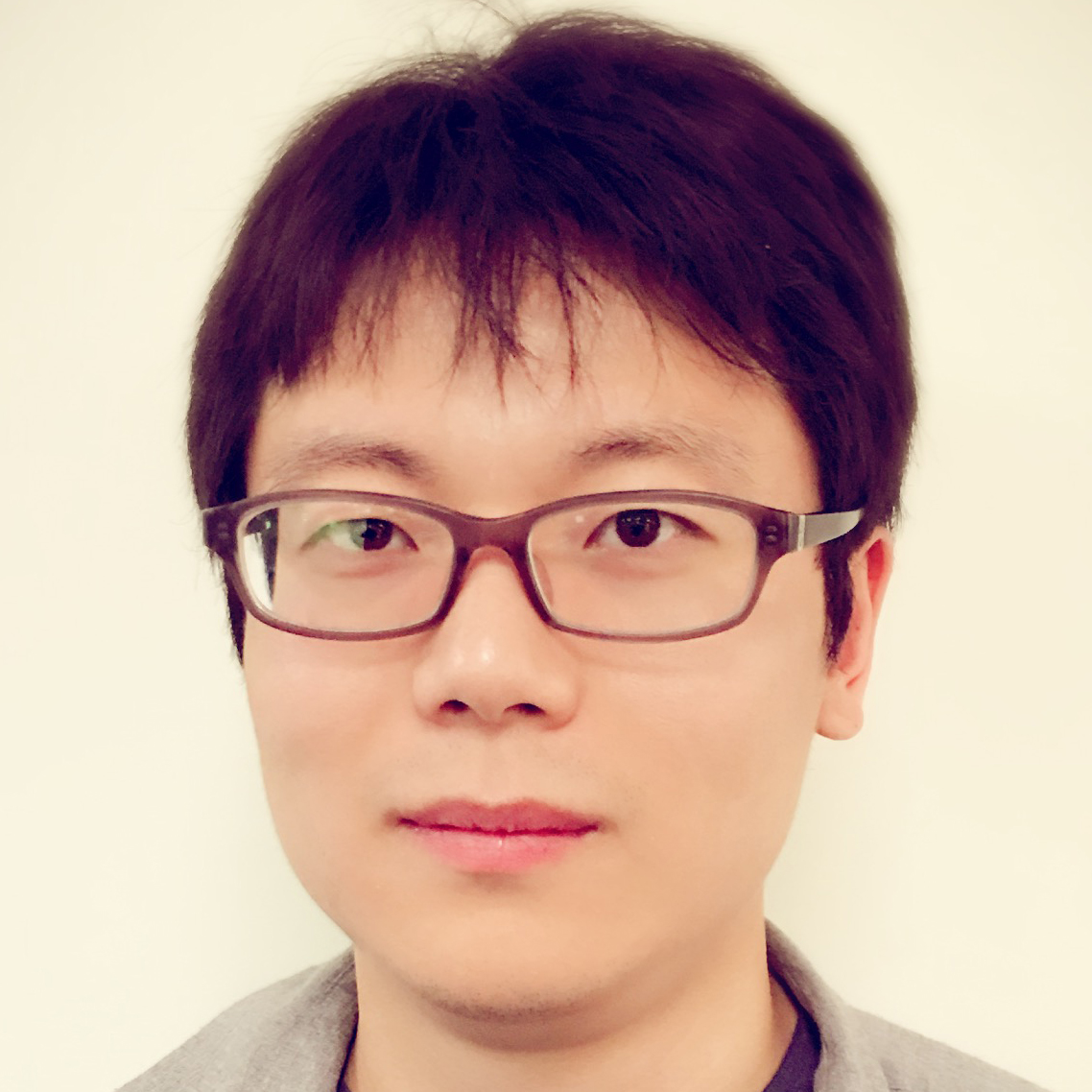}}]{Chao Wu}
is currently a Research Associate in Data Science Institute, Department of Computing, Imperial College London. His research is in the area of big data analysis, modelling and applications.
\end{IEEEbiography}

\begin{IEEEbiography}[{\includegraphics[width=1in,height=1.25in,clip,keepaspectratio]{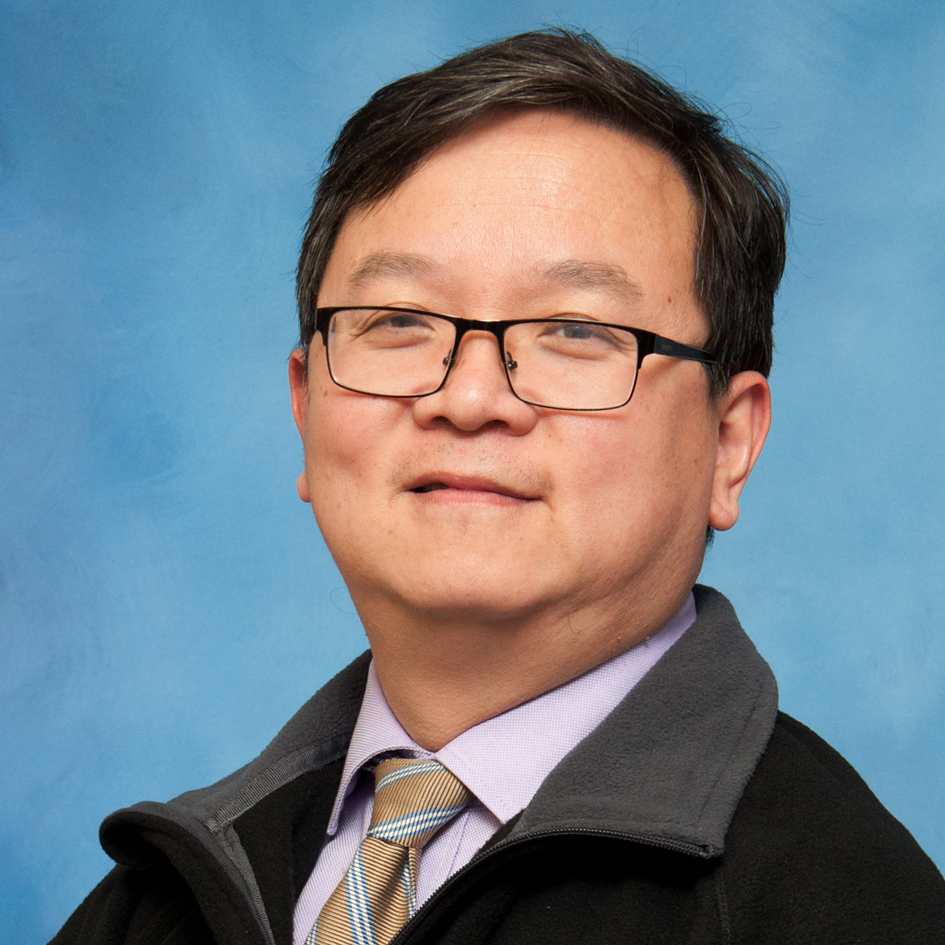}}]{Yike Guo}
is currently a Professor of Computing Science and a Director of Data Science Institute in the Department of Computing at Imperial College London. His research is in the areas of large scale scientific data analysis, data mining algorithms and applications, parallel algorithms and cloud computing.
\end{IEEEbiography}

\vfill







\end{document}